\pdfoutput=1

\documentclass[11pt]{article}

\usepackage[final]{acl}
\usepackage{xcolor}
\usepackage{graphicx}
\usepackage{colortbl}
\usepackage{adjustbox}
\usepackage{arydshln}
\usepackage{times}
\usepackage{booktabs}
\usepackage{multirow}
\usepackage{latexsym}
\usepackage[T1]{fontenc}
\usepackage{rotating} 
\usepackage{pifont}   
\usepackage[utf8]{inputenc}
\usepackage{microtype}

\usepackage{inconsolata}
\usepackage{tabularx}
\usepackage{graphicx}
\usepackage{subcaption} 
\usepackage{colortbl}
\usepackage{amsfonts}
\usepackage{amsmath}

\usepackage{authblk}
\usepackage[normalem]{ulem}
\usepackage{amsmath}
\title{Fast or Slow? Integrating Fast Intuition and Deliberate Thinking for Enhancing Visual Question Answering}
\author{
  \textbf{Songtao Jiang\textsuperscript{1}\textsuperscript{*}},
  \textbf{Chenyi Zhou\textsuperscript{1}\textsuperscript{*}}
  \textbf{Yan Zhang\textsuperscript{2}},
  \textbf{Yeying Jin\textsuperscript{3}},
  \textbf{Zuozhu Liu\textsuperscript{1,4†}}
\\
\\ 
  \textsuperscript{1}Zhejiang University,
  \textsuperscript{2}Byte Dance,
  \textsuperscript{3}National University of Singapore
\\
  \textsuperscript{4}Zhejiang Key Laboratory of Medical Imaging Artificial Intelligence
\\
  \small{
    \textbf{Correspondence\textsuperscript{†}:} \href{zuozhuliu@intl.zju.edu.cn}{zuozhuliu@intl.zju.edu.cn} 
  }
}
\begin{document}
\maketitle
\let\thefootnote\relax\footnote{* Equal contribution.}
\begin{abstract}

Multimodal large language models (MLLMs) still struggle with complex reasoning tasks in Visual Question Answering (VQA). While current methods have advanced by incorporating visual prompts, our study uncovers critical limitations: these approaches indiscriminately annotate all detected objects for every visual question, generating excessive visual markers that degrade task performance. This issue stems primarily from a lack of focus on key visual elements, raising two important questions: \textit{Are all objects equally important, and do all questions require visual prompts?} Motivated by {Dual Process Theory}, which distinguishes between instinctive and deliberate cognitive modes in human reasoning, we propose FOCUS, a plug-and-play approach that dynamically adapts to the complexity of questions, combining fast intuitive judgments with deliberate analytical reasoning to enhance the vision-language reasoning capability of the MLLM. For straightforward questions, FOCUS supports efficient zero-shot reasoning. For more complex tasks, it employs the {conceptualizing before observation} strategy to highlight critical elements. Extensive experiments on four benchmarks—ScienceQA, TextQA, VizWiz, and MME—demonstrate that FOCUS consistently improves the performance of both open-source and black-box MLLMs, achieving significant gains across all datasets. Ablation studies further validate the importance of combining diverse cognitive strategies with refined visual information for superior performance. Code will be released.

\end{abstract}

\section{Introduction}
Multimodal large language models (MLLMs) have demonstrated promising capabilities in visual question answering (VQA) tasks~\citep{OpenAI2023GPT4V,liu2023visual,chen2023minigptv2}. However, MLLMs still face challenges in addressing complex scenarios, particularly those requiring fine-grained visual perception and the ability to effectively leverage visual information for reasoning tasks~\citep{chen2024mllm,jiang2024joint}.

\begin{figure}
    \centering
    \includegraphics[width=1\linewidth]{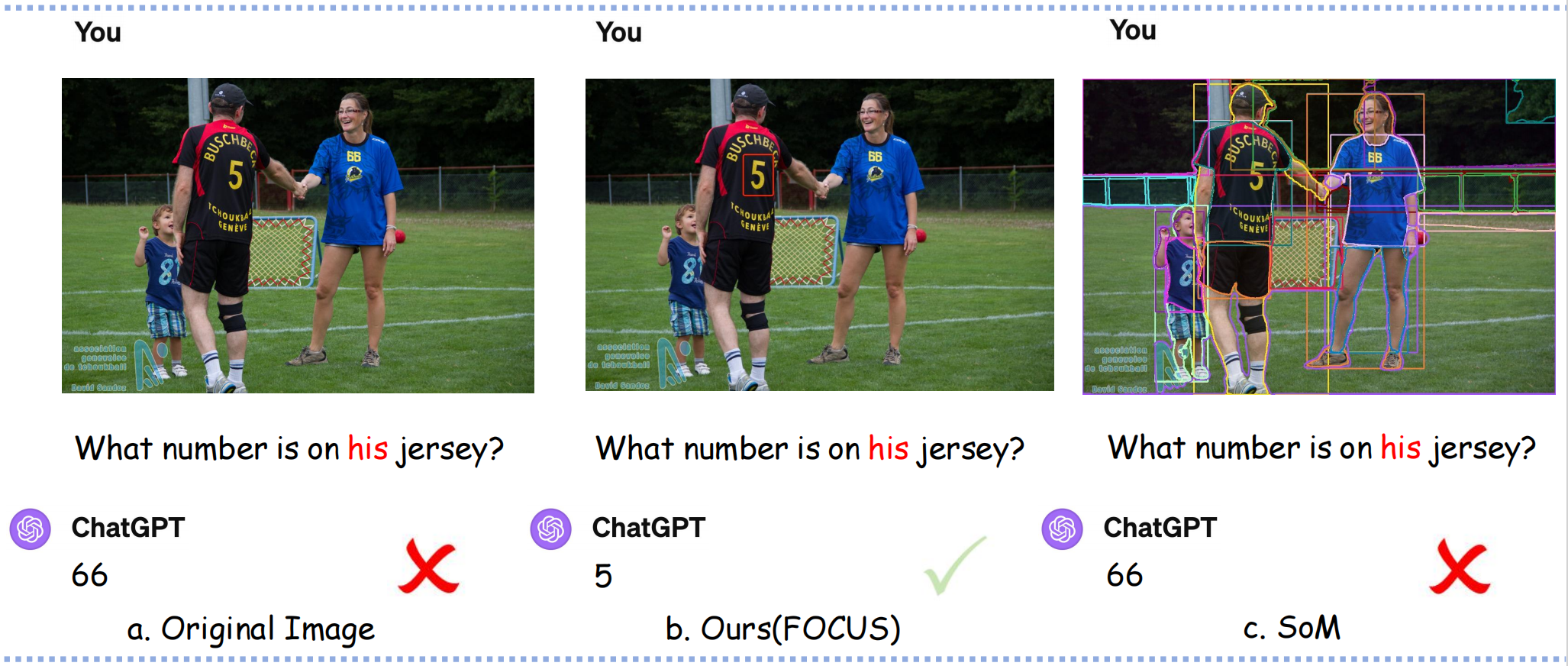}
\caption{Comparison of methods: baseline (original image), FOCUS, and SoM~\citep{yang2023set} (previous SoTA method). See Appendix~\ref{com_som} for more discussion.}
    \label{fig:example1}
\end{figure}

Current approaches~\citep{yang2023set,cai2024vip} employ segmentation models to generate visual prompts for all questions by annotating all objects in VQA images as visual markers, aiming to enhance MLLMs' attention to visual information for improving accuracy. While effective, as shown in Figure~\ref{fig:example1}, this coarse-grained guidance of adding visual markers to all objects fails to emphasize key elements. Consequently, MLLMs' attention is often distracted by redundant information, while also introducing significant computational overhead.

In this paper, we introduce FOCUS, a novel approach inspired by \textit{Dual Process Theory}~\citep{kahneman2011thinking}, which delineates two cognitive modes: a fast, instinctive mode (System 1) and a slower, deliberate mode (System 2). FOCUS enables MLLMs to dynamically alternate between rapid intuitive judgments and thorough analytical reasoning, adapting seamlessly to questions of varying complexity. Specifically, for each visual question, FOCUS first assesses its difficulty level. For simple questions, the MLLM performs efficient zero-shot reasoning. For more complex questions, FOCUS employs a \textit{conceptualizing before observation} strategy, which identifies and highlights critical visual elements using a segmentation model, enabling the MLLM to reason based on refined visual inputs.
Prior research has demonstrated that MLLMs struggle when visual elements in the input are ambiguous. As illustrated in Figure \ref{fig:example1}, FOCUS addresses this limitation by guiding MLLMs to focus on key visual elements, thereby unlocking their full reasoning potential. Furthermore, FOCUS not only enhances the accuracy of responses to complex questions but also optimizes computational efficiency, improving the overall performance.

We applied FOCUS to five MLLMs with diverse architectures and parameter scales, including three open-source models: LLaVA-1.5~\citep{liu2023improved}, MiniGPT4-V2~\citep{chen2023minigptv2}, and InstructBLIP~\citep{dai2023instructblip}, as well as two black-box models: GPT-4V~\citep{OpenAI2023GPT4V} and Gemini Pro~\citep{geminiteam2023gemini}. We evaluated their performance on four popular benchmarks, demonstrating that FOCUS consistently enhances model capabilities. Notably, FOCUS combined with LLaVA-1.5-13B achieves state-of-the-art (SoTA) performance across all four benchmarks, while FOCUS with LLaVA-1.5-7B delivers performance comparable to LLaVA-1.5-13B. Compared to the previous SoTA method, our approach not only surpasses SoM on all datasets but also reduces inference time by nearly 44\%. Additionally, extensive analysis and ablation studies validate the effectiveness of our dual-thinking strategy and demonstrate its capability to achieve attention calibration.

\begin{figure*}[ht!]
    \centering
    \includegraphics[width=0.9\linewidth]{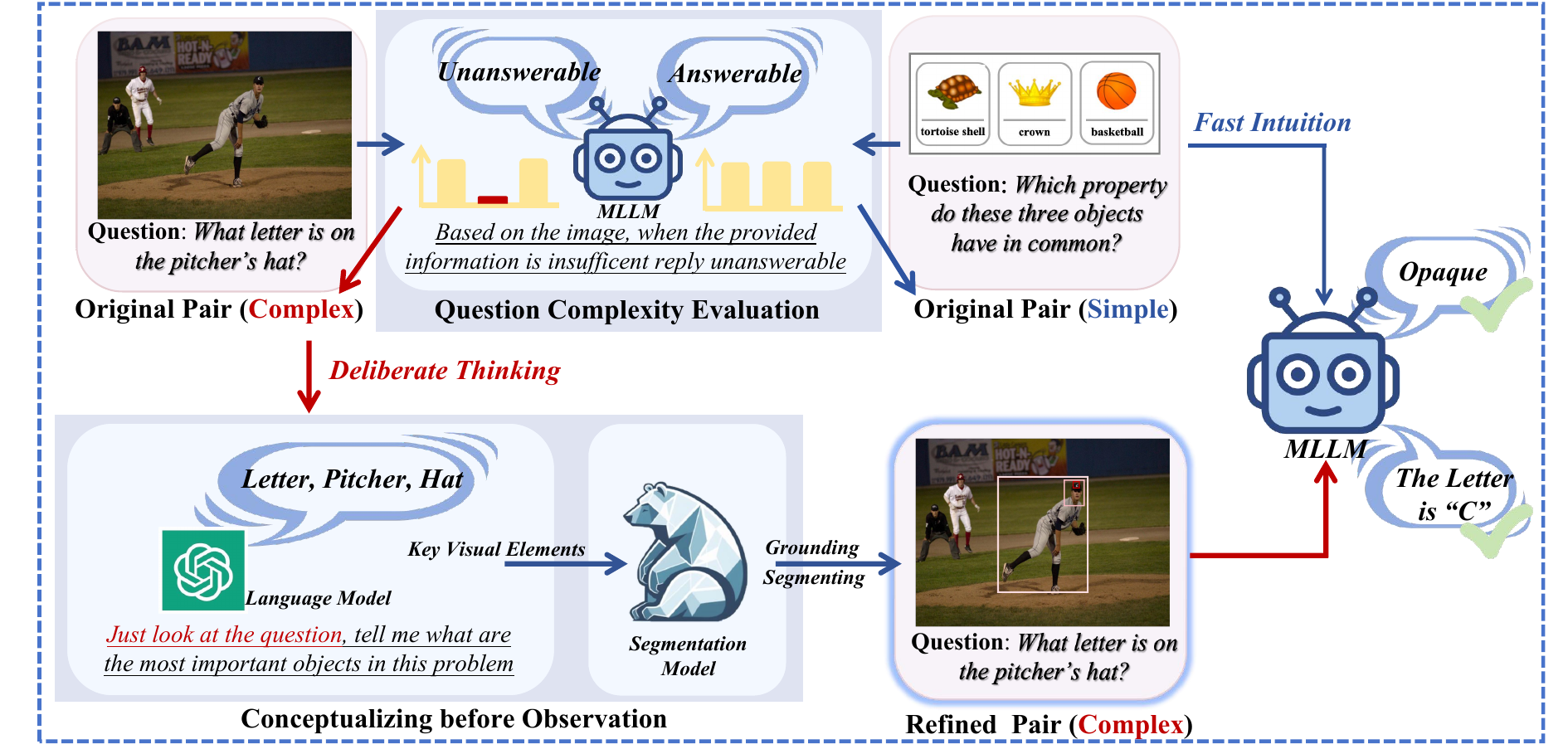}
    \caption{Overview of our model pipeline. (1) Question complexity evaluation for MLLM-based behavior to determine which questions require deliberate thinking. (2) Conceptualizing before observation, helping the model identify the most important visual information in the image.}
\label{fig:model}
\end{figure*}

\section{Methods}
\label{gen_inst}
\subsection{Problem Definition}
In this paper, we denote the dataset as $\mathcal{D}=\{(I_m, Q_m, A_m)\}^M_{m=1}$ where $M$ is the number of data samples. The goal of our tasks is to develop a mapping function $\mathcal{F}(\cdot)$ that can generate answers in response to the questions, represented as:
\begin{equation}
    A = \mathcal{F}(I, Q),
\end{equation}
Here, $I$ denotes the visual input. $Q$ represents the natural language question pertaining to the visual input $I$. $A$ is the output of the MLLM $\mathcal{F}(\cdot)$. FOCUS begins by evaluating question complexity to determine whether to apply the {conceptualizing before observation} strategy in the inference process. The overview of FOCUS is demonstrated in 
Figure \ref{fig:model}. The following sections will delve into these components in detail.

\subsection{Question Complexity Evaluation}
Past research \citep{xiong2023can} has shown that large language models effectively express their confidence in answers by mimicking human behavior, though they are prone to overconfidence. Building on this insight, we guide the MLLM to indicate their confidence before answering the question. If the MLLM demonstrates high confidence, we classify the question as simple. Otherwise, it is considered complex. Additionally, to mitigate the issue of overconfidence, we employ the self-consistency~\citep{wang2022self} method along with strict confidence evaluation criteria.

Specifically, we first set a high temperature for the MLLM and utilized the prompt strategy from VizWiz~\citep{gurari2018vizwiz} to activate the MLLM, generating $N$ responses $\{r_i\}^N_{i=1}$, where $r_i$ is either ``Answerable" or ``Unanswerable". A question is deemed simple if all responses are "Answerable." Otherwise, it is considered complex. In this paper, we set $N$ to three. We also discuss the effectiveness of the selected prompt and chosen \( N \) in Appendix~\ref{prompt}.

\subsection{Conceptualizing before Observation}
This strategy is specifically designed to address the complex question where the original visual information is confusing or insufficient. By conditioning the segmentation process on key extracted elements, it ensures that MLLMs can concentrate on the most crucial parts of the image corresponding to the question. 
To be specific, the first step involves extracting $K$ key elements $\{k_i\}^K_{i=1}$ from the question $Q$ using the language model.
Then the key elements $\{k_i\}^K_{i=1}$ are combined with bounding boxes generated by the segmentation model $\mathcal{S}$, which serves as an open-set object detector. The segmented regions $\{s_i\}^K_{i=1}$ are defined as:
\begin{equation}
s_i = \mathcal{S}(I, k_i)
\end{equation}
The final processed image $I'$ is produced by aggregating the relevant segments:
\begin{equation}
I' = \sum_{i \in K} s_i,
\end{equation}
Finally, the processed image $I'$, along with the original question $Q$, is then fed into the MLLM to generate a more accurate response. In this paper, we used GPT-3.5-turbo~\citep{openai_2023_gpt35turbo} as the language model and Grounded-SAM~\citep{ren2024grounded} as the segmentation model. The effectiveness of Grounded-SAM is discussed in Appendix~\ref{ablation_groudning}, while the effectiveness of GPT-3.5-turbo is discussed in Appendix~\ref{ablation_keywords}.

\begin{table*}[h!]
\centering
\small
\begin{tabularx}{\textwidth}{l l X X X l}                                   
\toprule
\textbf{Method} & \textbf{LLM} & \textbf{ScienceQA(\%)} & \textbf{TextVQA(\%)} & \textbf{VizWiz (\%)} & \textbf{MME} \\
\midrule
\rowcolor[HTML]{FFF9E3} BLIP-2~\citep{li2023blip}           & Vicuna-13B & 61.0    & 42.5  & 19.6 & 1293.8 \\
\rowcolor[HTML]{FFF9E3} IDEFICS-9B~\citep{laurenccon2024obelics}          & LLaMA-7B   & {---} & 25.9  & 35.5 & {---} \\
\rowcolor[HTML]{FFF9E3} IDEFICS-80B~\citep{laurenccon2024obelics}          & LLaMA-65B  & {---} & 30.9  & 36.0 & {---} \\
\rowcolor[HTML]{FFF9E3} Qwen-VL-Chat~\citep{bai2023qwen}        & Qwen-7B    & 68.2  & {61.5}  & 38.9 & 1487.5 \\
\rowcolor[HTML]{FFF9E3} LVIS-INSTRUCT4V~\citep{wang2023see}    & Vicuna-13B    & 69.0  & {62.1}  & 51.4 & 1572.0 \\
\midrule
\rowcolor[HTML]{E8F5E9} CCoT~\citep{mitra2024compositional}         & Vicuna-7B & {68.7}  &  58.9  & 51.1 & {---} \\
\rowcolor[HTML]{E8F5E9} DDCoT~\citep{zheng2023ddcot}         & Vicuna-7B & {67.6}  &  57.9  & 50.5 & {---} \\
\rowcolor[HTML]{E8F5E9} DCoT~\citep{jia2024dcot}       & Vicuna-7B & {69.3}  &  59.1  & 51.9 & {---} \\
\rowcolor[HTML]{E8F5E9} SoM~\citep{yang2023set}                &Vicuna-13B  & {71.3} & {61.5} & {54.0} & {1540.1}
\\
\rowcolor[HTML]{FCE4EC} {FOCUS + LlaVA-1.5-7B}  & Vicuna-7B & {70.2 {\color[HTML]{990000}}} & {60.1 {\color[HTML]{990000}}} & {54.5 {\color[HTML]{990000}}} & {1528.4 {\color[HTML]{990000}}} \\
\rowcolor[HTML]{FCE4EC} {FOCUS + LlaVA-1.5-13B} & Vicuna-13B & \textbf{74.4 {\color[HTML]{990000}(+3.1)}} & \textbf{63.6 {\color[HTML]{990000}(+2.1)}} & \textbf{58.5 {\color[HTML]{990000}(+4.5)}} & \textbf{1551.0 {\color[HTML]{990000}(+10.9)}}
\\
\bottomrule
\end{tabularx}
\caption{Comparison of methods across ScienceQA, TextVQA, VizWiz, and MME. Rows highlighted in \colorbox[HTML]{FFF9E3}{yellow} represent training-based methods, \colorbox[HTML]{E8F5E9}{green} represent inference-based methods, which are our primary comparison targets, and \colorbox[HTML]{FCE4EC}{pink} highlight FOCUS.}
\label{tab:main results}
\end{table*}

\begin{table*}[h!]
\centering
\small
\begin{tabularx}{\textwidth}{l l X X X l}                                   
\toprule
\textbf{Method} & \textbf{LLM} & \textbf{ScienceQA(\%)} & \textbf{TextVQA(\%)} & \textbf{VizWiz (\%)} & \textbf{MME} \\
\midrule
LLaVA-1.5         & Vicuna-7B & 66.8  & 58.2  & 50.0 & 1510.7 \\
LLaVA-1.5         & Vicuna-13B & 71.6  & 61.3  & 53.6 & 1531.3 \\
\rowcolor[HTML]{FCE4EC} 
LLaVA-1.5 + FOCUS & Vicuna-7B & \textbf{70.2 {\color[HTML]{990000}(+3.4)}} & \textbf{60.1 {\color[HTML]{990000}(+1.9)}} & \textbf{54.5 {\color[HTML]{990000}(+4.5)}} & \textbf{1528.4 {\color[HTML]{990000}(+17.7)}} \\
\rowcolor[HTML]{FCE4EC} 
LLaVA-1.5 + FOCUS & Vicuna-13B & \textbf{74.4 {\color[HTML]{990000}(+2.8)}} & \textbf{63.6 {\color[HTML]{990000}(+2.3)}} & \textbf{58.5 {\color[HTML]{990000}(+4.9)}} & \textbf{1551.0 {\color[HTML]{990000}(+19.7)}} \\
\midrule
MiniGPT4-V2         & LLaMA-7B & 56.7 & 34.1 & 44.1 & 1316.5 \\
\rowcolor[HTML]{FCE4EC} 
MiniGPT4-V2 + FOCUS & LLaMA-7B & \textbf{59.6 {\color[HTML]{990000}(+2.9)}} & \textbf{36.2 {\color[HTML]{990000}(+2.1)}} & \textbf{47.4 {\color[HTML]{990000}(+3.3)}} & \textbf{1334.3 {\color[HTML]{990000}(+17.8)}} \\
\midrule
InstructBLIP      & Vicuna-7B  & 60.5  & 50.1  & 34.5 & {1174.5} \\
\rowcolor[HTML]{FCE4EC} 
InstructBLIP + FOCUS & Vicuna-7B & \textbf{62.3 {\color[HTML]{990000}(+1.8)}} & \textbf{52.3 {\color[HTML]{990000}(+2.2)}} & \textbf{36.8 {\color[HTML]{990000}(+2.3)}} & \textbf{1189.8 {\color[HTML]{990000}(+15.3)}} \\
\bottomrule
\end{tabularx}
\caption{Comparative performance of open-source MLLMs across ScienceQA, TextVQA, VizWiz and MME. Both LLaVA-1.5 and MiniGPT-4-V2 use MLP as the connecting component between the ViT~\citep{dosovitskiy2020image} and LLM backbones, whereas InstructBLIP employs Q-Former~\citep{li2023blip}}
\label{tab:different MLLMs results}
\end{table*}

\section{Experiement}

\noindent\textbf{Datasets.} We evaluate FOCUS using three popular open-source MLLMs and two black-box MLLMs. The evaluation is conducted on four datasets specifically chosen to assess distinct capabilities: ScienceQA~\citep{lu2022learn}, TextVQA~\citep{singh2019towards}, VizWiz~\citep{gurari2018vizwiz}, and MME~\citep{fu2023mme}. For ScienceQA, TextVQA, and VizWiz, we measure accuracy, while for MME, we utilize the total score to evaluate overall performance. These datasets collectively address challenges in \textit{logical reasoning, text recognition, real-world visual understanding, and generalization}, offering a comprehensive assessment of FOCUS's capabilities. Additional details are in Appendix~\ref{app_datasets}.

\noindent\textbf{Baselines.} We evaluate two types of baselines for comparison. The first category comprises training-based approaches~\citep{Zhang2023Multimodal,li2023blip}, which leverage additional data or larger models. The second category includes inference-based methods, which utilize techniques such as cot reasoning or visual prompts~\citep{yang2023set,zheng2023ddcot,mitra2024compositional}.

\subsection{Main Results}
\noindent\textbf{Comparison with Previous Methods:} As shown in Table~\ref{tab:main results}, our method significantly outperforms previous approaches. Compared to the best-performing method, SoM, our approach not only achieves better performance but also offers a cost advantage, as it does not require adding unnecessary visual markers to the entire image.

\noindent\textbf{Applicability to Different MLLMs:} 
We evaluated FOCUS on multiple MLLMs, as detailed in Table~\ref{tab:different MLLMs results} and Table~\ref{tab:4v_experiment_comparison}. The results show that FOCUS consistently improves performance across diverse model architectures and parameter scales. For black-box MLLMs, we tested on a randomly selected subset of 250 samples from ScienceQA due to the cost.

\subsection{Ablation Study and Analysis}

\noindent\textbf{Ablation of Thinking Strategies.} As shown in Table~\ref{tab:self_prompting_results}, we assess the impact of Fast Intuition and Deliberate Thinking in FOCUS. Results reveal that uniformly applying a single strategy across all questions reduces performance. While Deliberate Thinking offers limited gains, it incurs significant computational costs. Conversely, relying solely on Fast Intuition hampers the MLLM's ability to address complex questions. Thus, adapting strategies based on question complexity is essential for optimal performance.

\noindent\textbf{How does Deliberate Thinking outperform SoM?} We compare the superiority of our Deliberate Thinking approach, which emphasizes {Conceptualizing before Observation}, with SoM's coarse-grained strategy of globally adding visual markers. As shown in Table~\ref{tab:dt_som}, our Deliberate Thinking method consistently outperforms SoM across all datasets, demonstrating its effectness in VQA.

\begin{figure}[ht]
    \centering
    \includegraphics[width=0.8\linewidth]{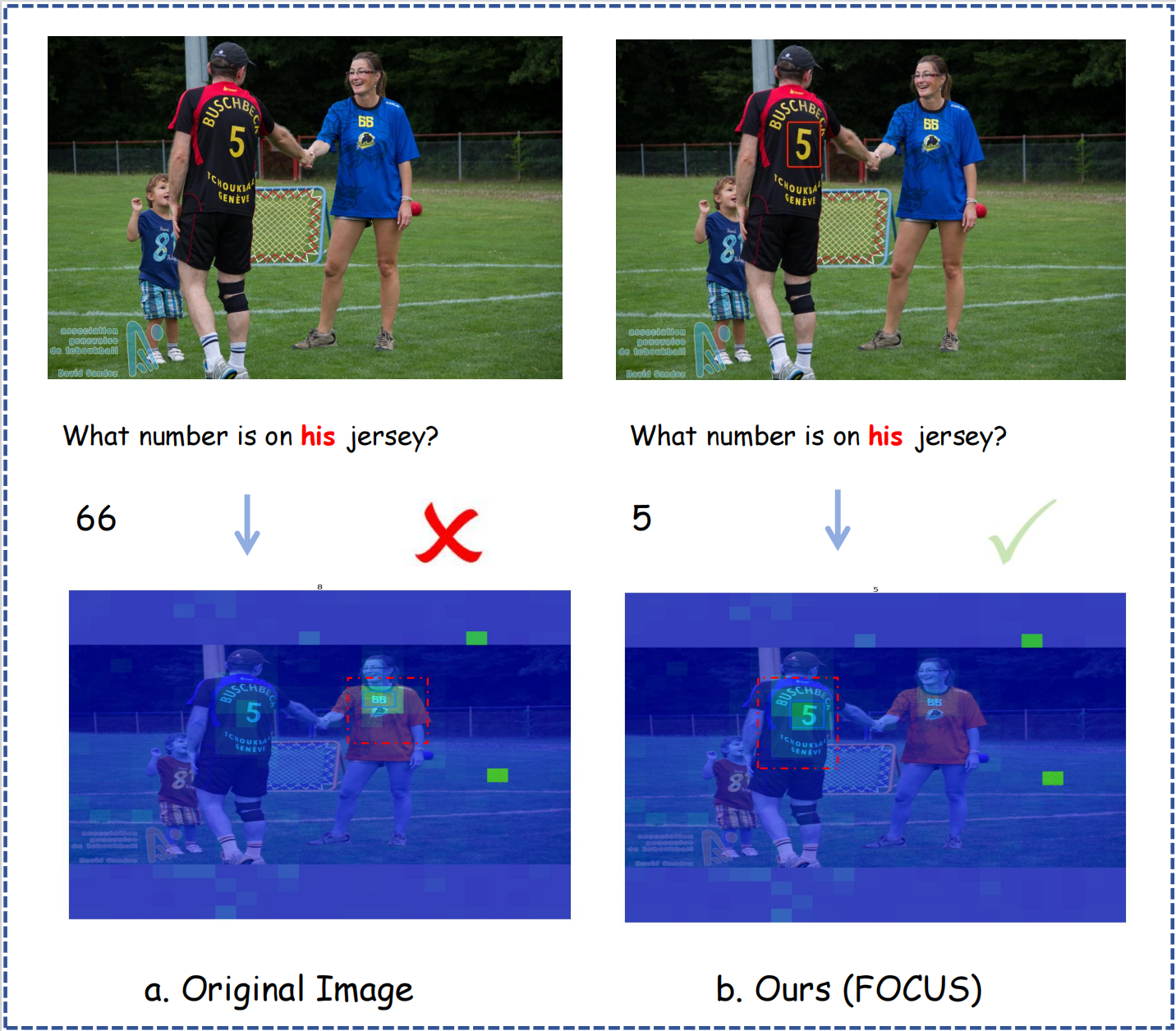}
    \caption{Visual attention visualization in LLaVA-1.5. The green areas indicate higher attention weights.}
    \label{fig:ana_case}
\end{figure}

\noindent\textbf{Visual Attention Calibration.} We assessed whether FOCUS's fine-grained visual markers enhance MLLMs' ability to filter distractions by analyzing the attention patterns of LLaVA-1.5. As shown in Figure \ref{fig:ana_case}, in the original image, the MLLM mistakenly focuses on the woman's jersey number, resulting in an incorrect answer. In contrast, FOCUS guides the MLLM's attention to the relevant image regions. This highlights FOCUS's ability to help MLLMs identify critical visual information, enhancing their perceptual accuracy.

\begin{figure}[ht]
    \centering
    \includegraphics[width=0.8\linewidth]{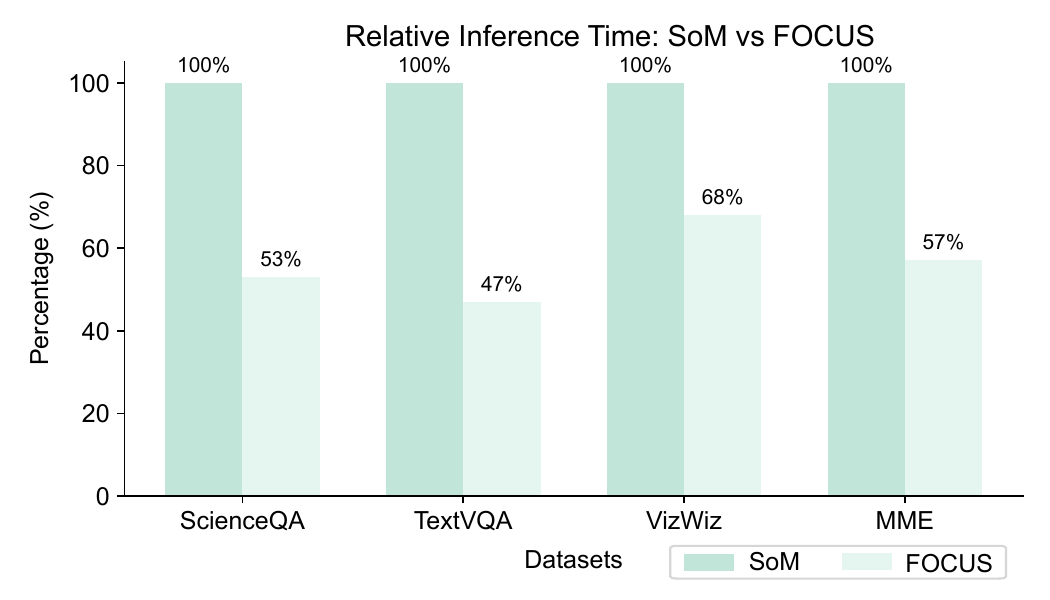}
    \caption{Relative inference time comparasion}
    \label{fig:cost}
\end{figure}
\noindent\textbf{Cost Analysis.} As illustrated in Figure~\ref{fig:cost}, our method achieves faster inference while maintaining superior performance compared to SoM. This efficiency stems from our approach's ability to skip the unnecessary addition of visual prompts for certain questions, as well as its avoidance of annotating all objects in the image—a process that is both time-consuming and detrimental to performance. These advantages demonstrate that our method is a cost-effective solution for VQA tasks.

\noindent\textbf{Analysis of Answerable and Unanswerable Proportions.}
We perform an analysis of the proportions of answerable versus unanswerable responses across different VLMs and benchmarks. Specifically, we compute average unanswerable proportions for three base VLMs, obtaining 39.00\% for LLaVA-1.6, 42.40\% for LLaVA-1.5, and 54.83\% for InstructBlip. These results indicate a clear trend: stronger models show lower unanswerable proportions, implying that more capable models have greater confidence and consequently prefer FI strategy. Additionally, we analyze average unanswerable proportions across four benchmarks—VizWiz, MME, TextVQA, and ScienceQA—and observe a ranking by decreasing uncertainty: VizWiz (58.33\%), MME (48.97\%), TextVQA (44.20\%), and ScienceQA (30.13\%). Our findings suggest that benchmarks demanding more detailed perceptual abilities elicit higher uncertainty, whereas simpler visual tasks, such as ScienceQA with cartoon-like images, enable models to respond with higher confidence.

\noindent\textbf{More Analysis of Unanswerable Questions.}
We further carefully analyze the unanswerable questions along with their associated images, identifying three prevalent scenarios where the model demonstrates increased uncertainty and thus necessitates DT strategy. First, images containing excessive elements, especially those with more than two primary objects, consistently confuse the models, irrespective of whether the style is realistic or cartoon-like (e.g., in ScienceQA). Second, images with unclear or ambiguous regions, notably common in the VizWiz benchmark, directly contribute to heightened uncertainty. Third, dense textual information in images, predominantly observed in TextVQA, visually distracts and overwhelms the VLMs, frequently causing confusion and errors. Additionally, we notice that some questions require specialized external knowledge, particularly within the ScienceQA benchmark, which involves physics or chemistry concepts. Although our FOCUS effectively directs model attention to relevant image regions, the absence of domain-specific knowledge occasionally leads to incorrect predictions. This limitation suggests a promising direction for future work, such as integrating Retrieval-Augmented Generation (RAG) techniques to enrich the model's knowledge base.

\noindent\textbf{Analysis of Confidence Evaluation Strategies.}
We conducted experiments to further investigate the impact of different confidence evaluation strategies. Specifically, we sampled 100 challenging questions from LLaVA-1.5, consisting of 50 questions incorrectly answered using DT but correctly with FI, and another 50 incorrectly answered using FI without employing DT. We compared three confidence measurement strategies: (1) our original approach (prompt with self-consistency), (2) semantic entropy computed from logits for binary classification of "answerable/unanswerable," and (3) self-consistency combined with semantic entropy.
The results are summarized in Table~\ref{tab:confidence_strategies}. Our analysis reveals several key observations: First, semantic entropy slightly improves performance by addressing the overly stringent requirement of consistent answerable predictions in the original method, reducing unnecessary DT usage on simpler questions. Second, incorporating self-consistency into semantic entropy further enhances performance by averaging entropy scores across multiple predictions, thereby reducing noise and improving robustness.

\section{Conclusion}
\label{sec:formatting}
We propose FOCUS, a novel approach that enhances MLLMs in VQA. By adapting thinking strategies to question complexity, FOCUS improves performance and resource efficiency. Experiments show it surpasses existing methods, offering a scalable solution for multimodal reasoning.

\section{Limitations}
While our FOCUS shows promise in enhancing MLLMs for vision-language tasks, it has several limitations. The method relies on external tools to provide fine-grained visual markers, which could introduce dependencies. Additionally, addressing the fundamental issues of visual hallucinations in MLLMs likely requires modifications at the model architecture and training levels, rather than solely relying on post-processing techniques. Therefore, exploring how to integrate this method more deeply with training processes presents an interesting direction for future research.
\section*{Acknowledgements}
This work is supported by the National Natural Science Foundation of China (Grant No. 12326612, 62476241), the Natural Science Foundation of Zhejiang Province, China (Grant No. LZ23F020008), and the Zhejiang University-Angelalign Inc. R\&D Center for Intelligent Healthcare.

\bibliography{anthology,custom}

\appendix

\section{Appendix}
\subsection{Comparison with SoM}
\label{com_som}
Although SoM~\citep{yang2023set} also employs visual prompts to enhance the performance of VLMs, our approach differs significantly in several key aspects:
\begin{enumerate}
    \item \textbf{Scope and Focus}: SoM primarily aims to unleash the visual grounding abilities of VLMs, with its experiments centered on tasks such as Open-vocabulary Image Segmentation, Referring Segmentation, Phrase Grounding, and Video Object Segmentation. These tasks are inherently focused on visual grounding rather than addressing challenges in the VQA domain. In contrast, we explore the potential of SoM in VQA and propose targeted improvements to adapt it for this domain.

    \item \textbf{Annotation Strategy}: SoM annotates all content in an image using a SAM-based model~\citep{kirillov2023segment}. However, our study reveals that this approach is less suitable for VQA tasks, as it fails to highlight critical information and may introduce coarse-grained visual markers that potentially disrupt the semantic integrity of the original image, thereby degrading VQA performance.

    \item \textbf{Prompting Efficiency}: Current visual or text prompting methods for VQA typically apply prompts uniformly to all questions. Our research demonstrates that not all questions require prompting, and we propose a more cost-efficient solution. This insight provides a new direction for the future development of visual prompting techniques.
\end{enumerate}

\section{More Analysis and Discussion}
\subsection{Effect of Chosen Prompt}
\label{prompt}
To evaluate whether our \textit{question complexity evaluation} strategy effectively identifies images that require further fine-grained processing, we used GPT-4V~\citep{OpenAI2023GPT4V}. We input a sample of 100 images from VizWiz~\citep{gurari2018vizwiz} along with their corresponding questions, using the prompt: "Is this VQA question a complex task requiring fine-grained perception, or a simpler task that only requires coarse-grained perception? Please provide a detailed explanation." After manually reviewing GPT-4V's reasoning, we labeled the images, determining that 59 required fine-grained visual information extraction, while 41 only needed coarse-grained perception.
We then explored the accuracy of question complexity evaluation with 1, 3, and 5 response iterations to assess the need for further fine-grained visual information extraction. The accuracies for each iteration were 59\%, 81\%, and 83\%, respectively. These results demonstrate that question complexity evaluation is highly effective in identifying which visual question answering (VQA) questions might confuse the model. This enables us to selectively apply fine-grained information marking to specific images, achieving satisfactory results with just three iterations.

\subsection{Effect of Segmentation Model}
\label{ablation_groudning}
We also evaluated the segmentation model's ability to mark the extracted keywords. Given that Grounded-SAM~\citep{ren2024grounded} was trained on a diverse set of open-world objects, and most VQA images come from the COCO dataset~\citep{lin2015microsoftcococommonobjects}, the marking accuracy is high. We randomly selected 100 samples from VizWiz, achieving an accuracy of over 85\%, confirming the effectiveness of our method. However, we set the threshold to 0.7 for the segmentation model. When the model is uncertain about the segmentation, it refrains from making further modifications to the image to avoid disrupting the original visual content. This ensures that the segmentation model does not negatively impact the MLLMs in cases of failure.

\subsection{Effect of Keywords Extraction}
\label{ablation_keywords}
We randomly selected 100 samples from VizWiz to assess GPT-3.5-Turbo's~\citep{openai_2023_gpt35turbo} ability to extract key concepts from VQA questions. Due to GPT-3.5-Turbo's strong information extraction capabilities~\citep{wei2023zero} and the relatively simple nature of the VQA questions, 87\% of the samples allowed for accurate keyword extraction. However, in the remaining 13\%, the relevant objects were not explicitly mentioned in the questions, preventing extraction. In such cases, the segmentation model refrains from adding markers, preserving the image's original semantic content.
\begin{table}[ht]
\centering 
\small
\begin{tabularx}{\columnwidth}{l X}
\toprule
\textbf{Method} & \textbf{ScienceQA(\%)} \\
\midrule
GPT-4V & 79.2 \\
Gemini Pro & 77.0 \\
\midrule
GPT-4V + FOCUS  & 82.4 {(+3.2)}  \\
Gemini Pro + FOCUS  & 79.4 {(+2.4)}  \\
\bottomrule
\end{tabularx}
\caption{Comparative performance of black-box MLLMs on a sample of 250 ScienceQA examples.}
\label{tab:4v_experiment_comparison}
\end{table}

\begin{table}[h!]
\centering
\small 
\begin{tabularx}{\columnwidth}{lXXXX} 
\toprule
\textbf{Method} & \textbf{ScienceQA (\%)} & \textbf{TextVQA (\%)} & \textbf{VizWiz (\%)}& \textbf{MME}\\
\midrule
FOCUS & 74.4 & 63.6 & 58.5 &1551.0 \\
Only FI. & 71.6  & 61.3  & 53.6 & 1531.3 \\
Only DT. & 72.2 & 61.6 &54.9 &1538.6 \\
\bottomrule
\end{tabularx}
\caption{Ablation study on Fast Intuition (FI) and DT (Deliberate Thinking). In this experiment, we used LLaVA-1.5 as the MLLM.}
\label{tab:self_prompting_results}
\end{table}

\begin{table}[ht!]
\centering
\small 
\begin{tabularx}{\columnwidth}{lXXXX} 
\toprule
\textbf{Method} & \textbf{SciQA. (\%)} & \textbf{TVQA. (\%)} & \textbf{VizWiz (\%)}& \textbf{MME}\\
\midrule
SoM  & {71.3} & {61.5} & {54.0} & {1540.1} \\
FI. + SoM & 72.6  & 61.8  & 55.2 & 1544.9 \\
\textbf{FOCUS (FI. + DT.)} & \textbf{74.4} & \textbf{63.6} & \textbf{58.5} &\textbf{1551.0} \\
\bottomrule
\end{tabularx}
\caption{Comparison of Deliberate Thinking (DT) and SoM. SciQA denotes ScienceQA, and TVQA denotes TextVQA.}
\label{tab:dt_som}
\end{table}

\subsection{Details of Datasets}
\label{app_datasets}
We evaluate FOCUS using three popular open-source MLLMs and two black-box MLLMs with diverse architectures: LLaVA-1.5~\citep{liu2023improved}, MiniGPT4-V2~\citep{chen2023minigptv2}, InstructBLIP~\citep{dai2023instructblip}, GPT-4V~\citep{OpenAI2023GPT4V}, and Gemini Pro~\citep{geminiteam2023gemini}. The evaluation is conducted on four datasets specifically chosen to assess distinct capabilities: 

\noindent\textbf{ScienceQA}~\citep{lu2022learn} primarily features images not from the real world, focusing on testing the logical reasoning capabilities of the models. We explore whether finer-grained visual information could enhance the models' reasoning abilities.

\noindent\textbf{TextVQA}~\citep{singh2019towards} contains images rich in challenging textual information, primarily evaluating the models' text recognition capabilities. We explore our model's ability to extract challenging fine-grained text information in complex visual scenes.

\noindent\textbf{VizWiz}~\citep{gurari2018vizwiz} consists mostly of real-life images, providing a comprehensive assessment of the MLLMs' proficiency in processing real-world imagery. This dataset includes images that contain complex visual elements, some of which are not very clear, making it difficult for models to interpret the content.

\noindent\textbf{MME}~\citep{fu2023mme} is a benchmark designed to evaluate MLLMs, addressing the lack of comprehensive assessment tools by measuring both perception and cognition across 14 diverse subtasks. We explore whether our approach could enhance the overall performance of MLLMs on this dataset.

For ScienceQA, TextVQA, and VizWiz, we assess accuracy, while for MME, we use the total score to evaluate the models' overall ability to understand and process different types of data.

\begin{table*}[t]
\centering
\resizebox{1.5\columnwidth}{!}{%
\begin{tabular}{lccc}
\toprule
\textbf{Strategy} & \textbf{DT (Count)} & \textbf{FI (Count)} & \textbf{Corrected (DT/FI)} \\
\midrule
Original Strategy (Prompt + Self-Consistency) & 50 & 50 & 0/0 \\
Semantic Entropy & 41 & 59 & 3/9 \\
Self-Consistency Entropy & 43 & 57 & 4/7 \\
\bottomrule
\end{tabular}%
}
\caption{Comparison of different confidence evaluation strategies on a subset of 100 challenging cases.}
\label{tab:confidence_strategies}
\end{table*}

\section{Related Work}
\noindent\textbf{Prompt Optimization in VQA}
Multimodal large language models (MLLMs) have shown promising capabilities in performing visual question answering (VQA) tasks~\citep{OpenAI2023GPT4V,XU2025126585,liu2023visual,chen2023minigptv2,jiang2024med,jiang2025omniv,jiang2024modality}.
Prior research aimed at improving MLLMs for Visual Question Answering (VQA) has primarily focused on gradient-based methods~\cite{li2021prefix,vu2021spot,gu2021ppt,liu2023gpt,mokady2021clipcap,qian2022controllable,an2022input} and prompt optimization techniques~\cite{deng2022rlprompt,sun2023offline,jiang2024joint,Zhang2023Multimodal}. Among these, the Multimodal Chain of Thought (MM-CoT) method~\cite{Zhang2023Multimodal} is particularly notable for its ability to integrate visual and textual information within LLMs, achieving superior performance in reasoning tasks. However, this approach incurs higher training costs.

Recently, visual prompt strategies have gained prominence in MLLMs. The Set-of-Mark (SoM) method~\cite{yang2023set} is the first to experiment with applying visual marks as prompts on image inputs for MLLMs like GPT-4V, aiming to enhance their grounding capabilities. SoM has demonstrated state-of-the-art results on datasets such as RefCOCOg and DAVIS2017~\cite{kazemzade2014referring,openai2023gpt}. However, its application to VQA tasks remains underexplored. Moreover, we observe that SoM often segments the entire image when applied to VQA tasks, which may overlook critical information necessary for accurate answers. In contrast, our approach introduces more precise segmentation in MLLMs, focusing on the most relevant parts of the image to significantly improve VQA performance.

\subsection{Case Study}
There are more case studies on GPT-4V:
\begin{figure*}[ht!]
    \centering
    \includegraphics[width=0.8\linewidth]{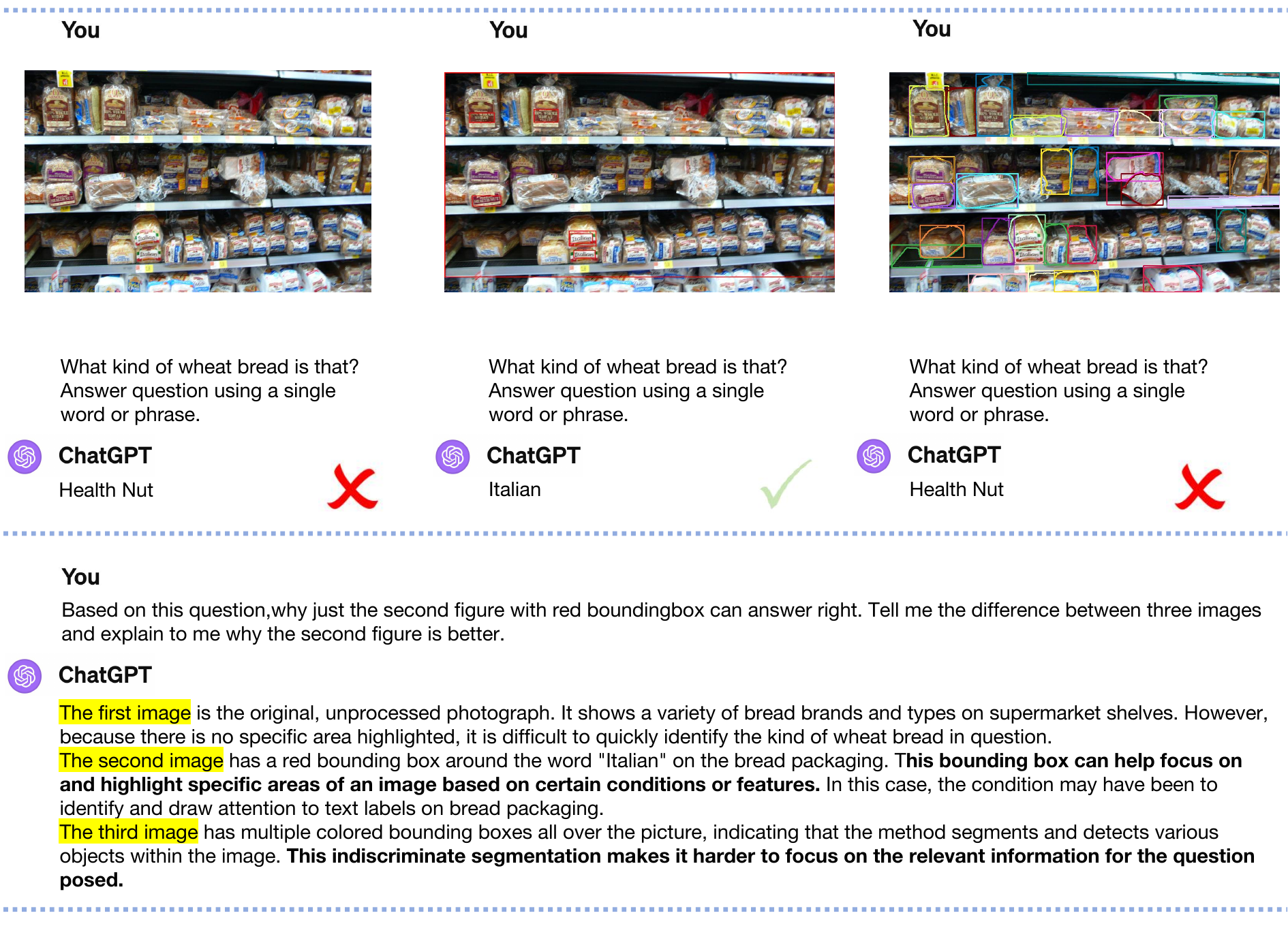}
    \caption{More cases on GPT-4V}
    \label{case_study_appen1}
\end{figure*}

\begin{figure*}[ht!]
    \centering
    \includegraphics[width=0.8\linewidth]{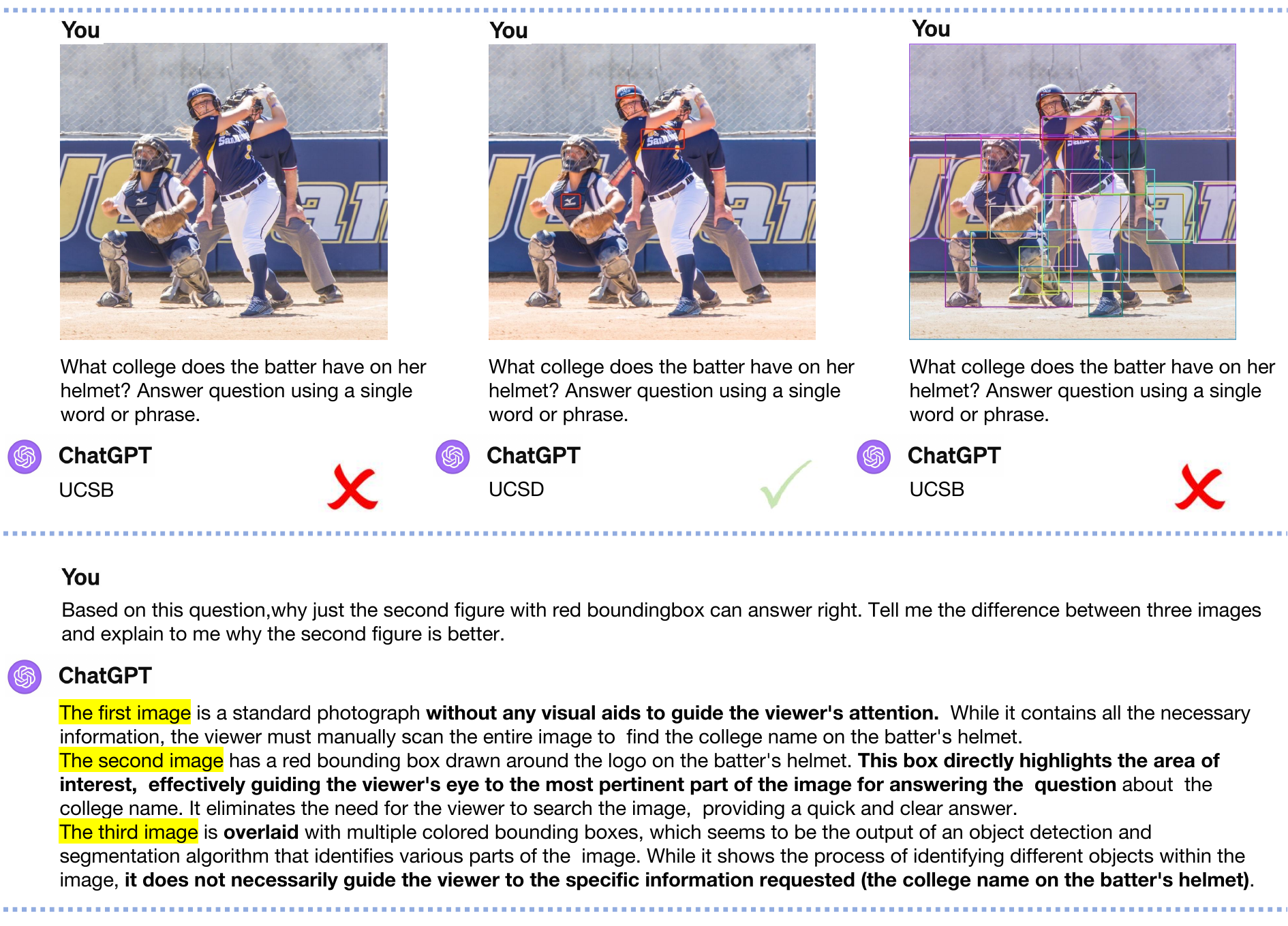}
    \caption{More cases on GPT-4V}
    \label{case_study_appen2}
\end{figure*}
\end{document}